\documentclass[letterpaper, 10 pt, conference]{ieeeconf}  % Comment this line out if you need a4paper

\usepackage{hhline}
\usepackage{dingbat}
\usepackage{multirow}
\usepackage{multicol}
\newcommand{\mbf}[1]{{\mathbf{#1}}}
\newcommand{\refsec}[1]{Sec.~\ref{sec:#1}}
\newcommand{\reffig}[1]{Fig.~\ref{fig:#1}}
\newcommand{\refeq}[1]{Eq.~\ref{eq:#1}}
\newcommand{\reftab}[1]{Table~\ref{tab:#1}}

\newcommand{\ie}[1][ ]{{\em i.\thinspace{}e\@.{}}#1}
\newcommand{\eg}[1][ ]{{\em e.\thinspace{}g\@.{}}#1}
\newcommand{\ea}{{\em et al.}}
\newcommand{\argmin}{\operatorname*{argmin}}

\newcommand{\pose}[2]{^{#1}\mathbf{T}_{#2}}
\newcommand{\img}[1]{\mathbf{I}_{#1}}
\newcommand{\pt}{\mathbf{p}}
\usepackage{amsmath,bm}
\usepackage{amsfonts}
\usepackage{graphicx}
\IEEEoverridecommandlockouts                              % This command is only needed if 
\title{\LARGE \bf
Instance-aware multi-object self-supervision for \\ monocular depth prediction
}

\author{Houssem eddine BOULAHBAL$^{1}$, Adrian VOICILA$^{2}$ and Andrew I. COMPORT$^{3}$% <-this % stops a space
\thanks{*This work was performed using HPC resources from GENCI-IDRIS (Grant 2021-011011931). The authors would like to acknowledge the Association Nationale Recherche Technologie (ANRT) for CIFRE funding (n°2019/1649).}% <-this % stops a space
\thanks{$^{1}$H. BOULAHBAL is with Renault Software Factory and CNRS-I3S, Côte d’Azur University
        2600 Rte des Cretes, 06560 Valbonne, France and
        2000 Route des Lucioles BP 121, Sophia Antipolis, France
       {\tt\small Houssem-eddine.Boulahbal@renault.com}}%
\thanks{$^{2}$A. VOICILA is with  Renault Software Factory
        2600 Rte des Crêtes, 06560 Valbonne, France 
        {\tt\small Adrian.Voicila@renault.com}}%
\thanks{$^{3}$A.I. COMPORT is with CNRS-I3S, Côte d’Azur University
2000 Route des Lucioles BP 121, Sophia Antipolis, France
{\tt\small Andrew.Comport@cnrs.fr}}%
}

\begin{document}

\maketitle
\thispagestyle{empty}
\pagestyle{empty}

\begin{abstract}
This paper proposes a self-supervised monocular image-to-depth prediction framework that is trained with an end-to-end photometric loss that handles not only $6-$DOF camera motion but also $6-$DOF moving object instances. Self-supervision is performed by warping the images across a video sequence using depth and scene motion including object instances. One novelty of the proposed method is the use of the multi-head attention of the transformer network that matches moving objects across time and models their interaction and dynamics. This enables accurate and robust pose estimation for each object instance. Most image-to-depth predication frameworks make the assumption of rigid scenes, which largely degrades their performance with respect to dynamic objects. Only a few SOTA papers have accounted for dynamic objects. The proposed method is shown to outperform these methods on standard benchmarks and the impact of the dynamic motion on these benchmarks is exposed. Furthermore, the proposed image-to-depth prediction framework is also shown to be competitive with SOTA video-to-depth prediction frameworks.
\end{abstract}
\section{Introduction}
\begin{figure}
    \centering
    \includegraphics[width=0.5\textwidth]{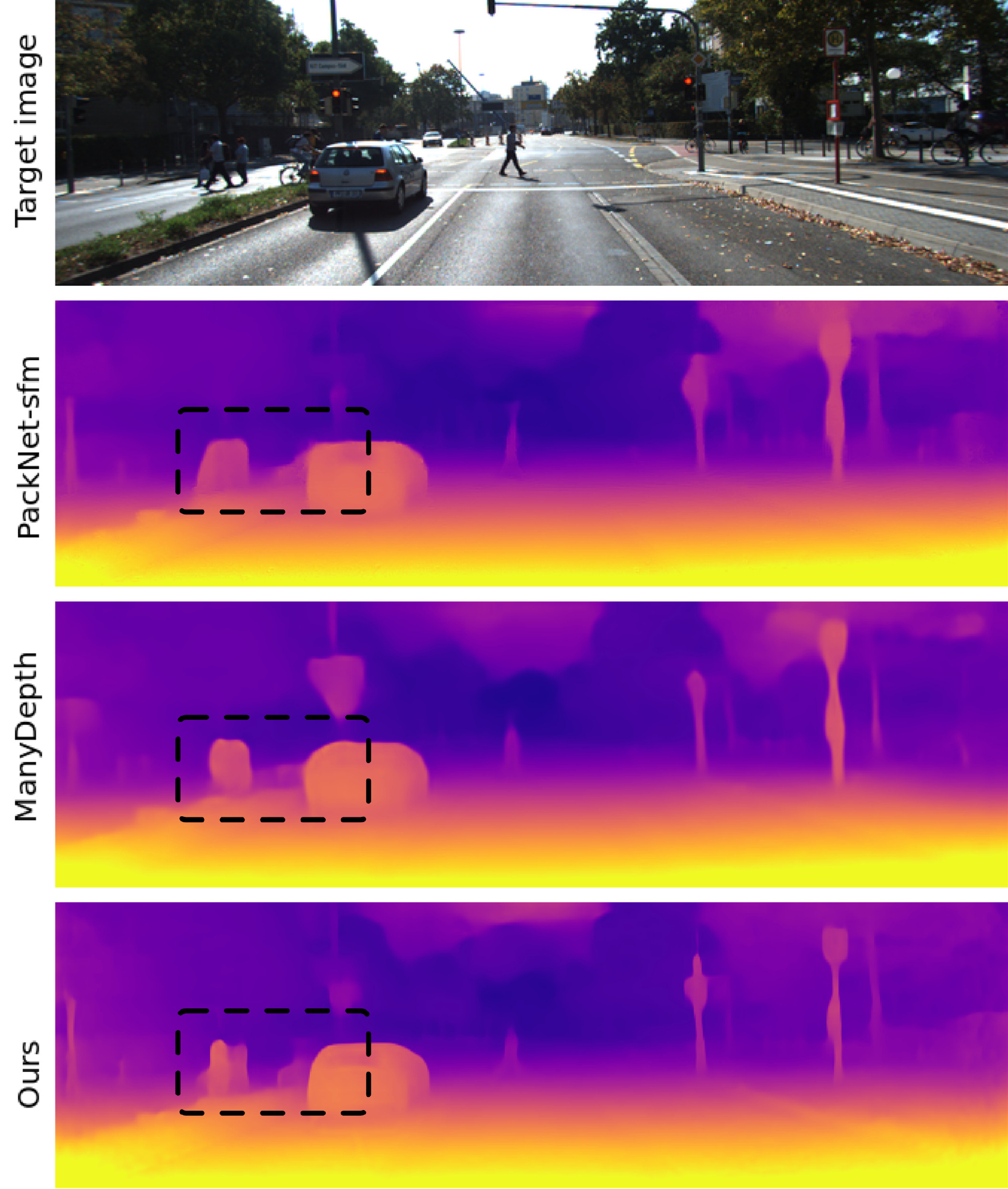}
    \caption{Qualitative results of the proposed method with SOTA Methods~\cite{Watson2021,Rares2020}. The proposed method produce a high quality depth and it is able to account for dynamic and small object in the scene. Comapred to the other methods, this example shows that the proposed method is able to handle better the dynamic objects \ie the pedestrian and the bicycle. This observation is validated with quantitative results in \refsec{results}}
    \label{fig:qualitative}
\end{figure}

\begin{figure*}
    \centering
    \includegraphics[width=\textwidth]{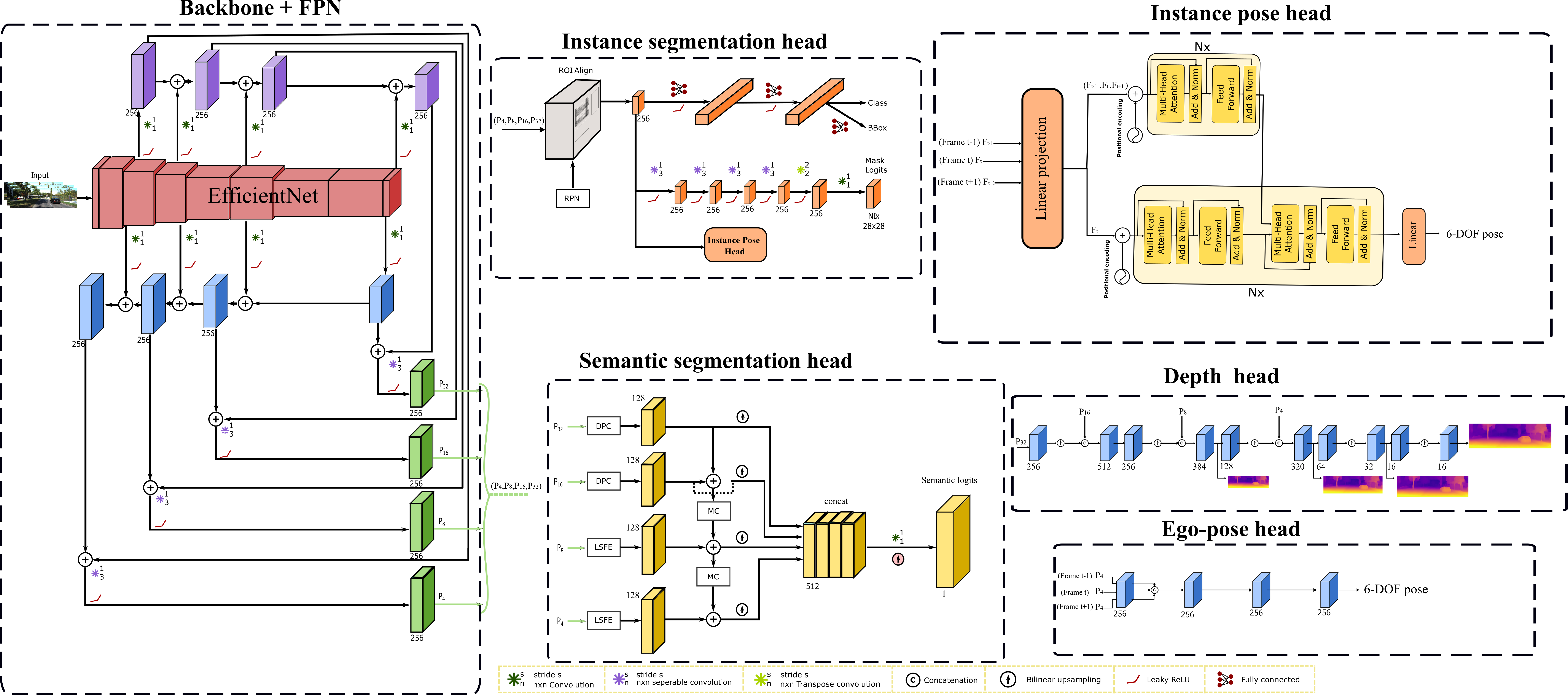}
    \caption{The proposed model architecture consisting of the EfficientNet backbone~\cite{tan2019efficientnet}, BiFPN~\cite{tan2020efficientdet}, the DPC~\cite{chen2018searching} semantic head, the MaskRCNN instance segmentation head~\cite{he2017mask}, the novel instance pose head, an ego-pose head and a depth head. During training, the FPN features ($P_4, P_8, P_{16}, P_{32}$) are extracted for the source $\img{t}$ and target frames $\img{t-1}$, $\img{t+1}$. These features are pooled using the proposals of the RPN and the ROI Align modules. The class, bounding box and instance mask heads use only the features of frame $\img{t}$. The Instance pose head uses both source and target frames as input. This head output a $6$ axis-angle parameters for each instance. Similarly, the ego-pose head uses the both source and target frames $P_4$ FPN' features as input. This head output a $6$ axis-angle parameters for the ego-pose. The depth head input the FPN features of the source frame $\img{t}$ and output a multi-scale depth.}
    \label{fig:architecture}
\end{figure*}
Monocular depth prediction is a prominent problem in computer vision, with many applications in robotics, AR/VR, autonomous driving and its related downstream tasks, \eg automatic emergency braking (AEB). While Lidar scans provide high accuracy depth measurements, generating high quality depth from a monocular camera is attractive due to the cost-efficiency and availability of such systems. Recent developments based on deep learning methods~\cite{Zhou2017,Gordon2019,Shu2020,Bian2019,Chen2019,Rares2020,Wang2021,Watson2021,chen2019self,Ranjan2019,luo2019every} have demonstrated competitive depth prediction quality. These methods are formulated via {\em self-supervised learning} based on making use of monocular video during training. Thus, large and accurate ground-truth datasets are not required. One way to design such a training model is to introduce a depth network alongside a pose network that can estimate the pose between two video frames. These networks are optimized using a photometric loss between the target image and a set of warped images obtained using the pose and the depth.

% The problem : violation of the rigid scene assumption
A common assumption for self-supervised training is the rigid scene assumption,~\ie a static scene and a moving camera. However, this assumption is often violated due to the motion of moving objects present in the scene. A possible solution is to mask the pixels of dynamic objects using either a learned mask~\cite{Zhou2017}, semantic guidance~\cite{Klingner2020} or auto-masking~\cite{Godard2019,Watson2021}. However, these solutions are merely a workaround to avoid non rigid-scenes and they subsequently miss data on moving objects that could otherwise be useful for further constraining depth prediction.~\cite{Ranjan2019,Yin2018, Vijayanarasimhan,lee2021learning,xu2021moving} have addressed the moving object in the scene.~\cite{Vijayanarasimhan,lee2021learning,xu2021moving} learn a per-object semantic segmentation mask and a motion field that accounts for the moving object.~\cite{Ranjan2019,Yin2018} rely on the optical flow which does not explicitly model the $6-$DOF motion of objects separately. These networks are optimized for local rigidity and the notion of the class and the possible dynamics for each class is not taken into account.
%Alternatively, the proposed network explicitly detects and segments the moving objects enabling global consistency, class dynamics and $6$-DOF pose for each instance. The proposed method demonstrates high quality depth estimation and state-of-the-art results on the KITTI benchmark.

% The proposition explained
A proposition is made in this paper to alleviate this assumption. Non-rigid scenes are learnt by factorizing the motion into the dominant ego-pose and \textbf{a piece-wise rigid pose for each dynamic object} explicitly. Therefore, for static objects, only the ego-pose is used for the warping, while the dynamic objects are subject to two transformations using the motion of the camera and the motion of each moving object. The proposed method explicitly models the motion of each object allowing accurate warping of the scene elements. In order to model this motion,
The proposed method makes use of the multi-head attention of the transformer network that matches moving objects across time and models their interaction and dynamics. This enables accurate and robust pose estimation for each object instance. The proposed method achieves SOTA results on the KITTI benchmark. In summary, the contributions of this paper are:

% Talk about KITTI being predominated with static 86.43%

%a pretrained instance segmentation network is modified by adding a $6$-DOF ego-pose branch and a %TRM pose head to each instance \reffig{architecture}. These heads leverage a transformer encoder-decoder~\cite{Vaswani} in order to match each object in the consecutive frame in the video during training and learn a $6$-DOF pose for each object. Using the mask of the instance, the pixels of each dynamic object instance are subject to the aforementioned pose transformations.

%Concurrent works~\cite{Ranjan2019,Yin2018} rely on the optical flow which does not explicitly model the $6-$DOF motion of objects separately. These networks are optimized for local-rigidity and the notion of the class and the possible dynamics for each class is not taken into account.~\cite{Vijayanarasimhan,lee2021learning,xu2021moving} learn a per-object semantic segmentation mask and a motion field. Alternatively, the proposed network explicitly detects and segments the moving objects enabling global consistency, class dynamics and $6$-DOF pose for each instance. The proposed method demonstrates high quality depth estimation and state-of-the-art results on the KITTI benchmark.

\begin{itemize}
    \item A novel network architecture based on the transformers multi-head attention that explicitly models the dynamics of moving objects. 
    \item An accurate and robust per-object pose is obtained by matching and modeling the interaction of the objects across time. 
    \item High quality depth prediction achieving competitive performance with respect to state-of-the-art results on the KITTI benchmark~\cite{Geiger2012CVPR}. 
    \item The demonstration that the KITTI benchmark has a bias favoring static scenes and a method to test the quality of moving object depth prediction.
\end{itemize}

% Name the contribution

% The assumption of monocular static scene and moving camera
% The proposed Instance pose warping
% Small conclusion about the contribution of the paper
\section{Related work}
\subsection{Self-supervised depth prediction}
Depth prediction has been successful with self-supervised learning from videos. The seminal work of Zhou \ea~\cite{Zhou2017} introduced the core idea to jointly optimize the pose and depth network using image reconstruction and photometric loss. To account for the ill-posed nature of this task, several works have addressed different challenges.~\cite{Vijayanarasimhan,xu2021moving,luo2019every} addressed the rigid scene assumption.~\cite{Gordon2019,Shu2020} proposed more robust image reconstruction losses to reject the outliers such as occlusion during training.~\cite{chen2019self} have addressed learning the camera parameters for better generalization.~\cite{Bian2019,Chen2019,Rares2020,Wang2021} have addressed the scale ambiguity problem and propose to enforce depth scale and structure consistency.~\cite{mccraith2020monocular, Watson2021} employed a test-time refinement by allowing the model parameters to vary dynamically during inference using a photometric loss.
Similarly, the proposed paper jointly optimize the pose and depth using the photometric loss as supervisory signal and addresses specifically the problem of moving objects in the scene.   
%~\cite{Klingner2020,Yin2018,Ranjan2019,Chen2019}

\subsection{Camera and object motion factorization}
Supervising the depth with a photometric loss is problematic when moving objects are present in the scene. This challenge has gained attention in the literature. A common solution is to disentangle the dominant ego-motion and the object motion.~\cite{chen2019self,Ranjan2019,Yin2018,hur2020self} leverage an optical flow network to detect moving objects by comparing the optical flow with depth-based mapping.
~\cite{lee2021attentive} learns a monocular depth in order to estimate the motion field as two stage learning.~\cite{Vijayanarasimhan} learns a per-object semantic segmentation mask and a motion field is obtained by factorization of the motion of each mask and the ego-motion.~\cite{safadoust2021self} addresses the object motion without additional labels by proposing a scene decomposition into a fixed
number of components where a pose is predicted for each component.~\cite{xu2021moving} relaxes the problem using local rigidity within a predefined window.~\cite{luo2019every} leverages the geometric consistency of depth, ego-pose and optical flow and categorises each pixel as either rigid motion, non-rigid/object motion or occluded/non-visible regions.
A recent work that is closest to the proposed method is Insta-DM~\cite{lee2021learning}. In that method, the source and target images are masked with semantic masks and an object PoseNet is used to learn the pose from the masked RGB images. Alternatively, the method proposed in this paper factorizes the motion into ego-motion and object-motion and exploits a transformer attention network to perform instance segmentation and learn a per-object motion.

\section{Method}
\subsection{Problem formulation}
The aim of monocular depth prediction is to learn an accurate depth map through the mapping $\mbf{D}_t = f(\img{t}; \bm{\theta})$ where $\img{t} \in \mathbb{R}^{W\times H\times3}$ the target image and $\mbf{D}_t \in \mathbb{R}^{W\times H\times1}$ target depth. 
%and $\bm{\theta}$ is the network parameters. 
In self-supervised learning, this model is trained via novel view synthesis by warping a set of source frames $\mbf{I}_{s}$ to the target frame $\mbf{I}_{t}$ using the learned depth $\mbf{D}_{t}$ and the target to source pose $\pose{s}{t} \in \mathbb{SE}[3]$. Prior methods assume a static scene observed by a camera undergoing ego-motion. This fundamental assumption is often violated when moving objects are present in the scene. A common solution is to mask the dynamic objects' pixels. These solutions aim to assist the static scene assumption. In this paper, rather than enforcing the rigid scene restriction, a proposition is made to alleviate this restriction. For each pixel, a \textbf{global rigid-scene pose} and a \textbf{piece-wise rigid pose} for each dynamic object is learned. This is more precise and consistent with the non-rigid real-world situations. An instance segmentation network~\cite{mohan2021efficientps} is extended to incorporate the pose information so that the network learns an additional $6$-DOF pose for each instance. Therefore, each instance $i$ is represented by the class $c^i$, bounding box $\mbf{B}^i$, mask $\bm{\mathcal{M}}^i$ and the additional pose $\pose{}{o}^i \in \mathbb{SE}[3]$ as illustrated in \reffig{architecture}. The per-instance warping is defined as:
%NOTATION The per-instance 
\begin{equation}\footnotesize\label{eq:warping}
        \widehat{\pt}_{s} \sim \mbf{K}  \sum_{i=0}^{m} \big[ \mathcal{M}^i_{\pt_{t}} \mbf{T}_{o}^i  + (1-\bm{\mathcal{M}}^i_{\pt_{t}}) \textit{\textbf{I}}_4 \big] \pose{s}{t} \mbf{D}_{t}  \mbf{K}^{-1} \pt_{t}
\end{equation}
where $m$ is the number of dynamic object instances and $\textit{\textbf{I}}_4$ a $4\times 4$ identity matrix. For simplicity the homogeneous pose and projection transformations are omitted in~\refeq{warping}. The mask $\mathcal{M}^i$ is used to transform only the dynamic object $i$ with its pose $\mbf{T}_{o}^i$. Rigid scene points are transformed only with the pose $\pose{s}{t}$. Using the \refeq{warping}, the image $\widehat{\mbf{I}}_t$ is obtained by inverse warping.
% The model explicitly models the motion for each object. Therefore, all the pixels are used for training the model.  

\subsection{Architecture}
% Quick overview of the architecture
In order to explicitly model the motion of the moving objects, an instance pose head is introduced into an instance segmentation network. EfficientPS~\cite{mohan2021efficientps} has demonstrated SOTA results for panoptic and instance segmentation and is therefore adopted in this paper for depth prediction.
It consists of the EfficientNet backbone~\cite{tan2019efficientnet}, BiFPN~\cite{tan2020efficientdet}, MaskRCNN instance segmentation head~\cite{he2017mask} and the DPC~\cite{chen2018searching} semantic head.
The EfficientNet backbone has demonstrated its success as a task agnostic feature extractor for nearly all vision tasks. It is easily scalable allowing more complexity/FLOPS trade-off. The BiFPN allows low-level and high-level feature aggregation thus, enabling a rich representation that accounts for the fine-details and more global abstraction at each feature map. During training, the FPN features ($P_4, P_8, P_{16}, P_{32}$) are extracted for the source and target frames. The two pose heads use both source and target features while the instance, semantic and depth heads use only the target features. The model architecture is shown in~\reffig{architecture}. The additional heads are detailed in the following. 

\subsubsection{Instance pose head}  
% Small intro
The key idea of this paper is to factorize the motion by explicitly estimating the $6-$DOF pose of each object in addition to the dominant ego-pose. In order to accurately estimate this motion, the objects should be matched and tracked temporally and its interaction should be modeled. Inspired by the prior work on object tracking~\cite{meinhardt2021trackformer,xu2021transcenter}, a novel instance pose head that extends the instance segmentation is proposed using transformer module~\cite{Vaswani}. This head makes use of the multi-head attention to learn the association and interaction of the object across time.

% where does the features come from
The RPN network yields N proposals. The features of each proposal are pooled using a ROI Align module. These features are extracted for the three frames. Therefore, the input of the instance pose head is $b \times (s+1) \times N \times 256 \times 14\times 14$. Where $b$ is the batch size and $s$ is the number of sources images. 
% Linear projection
The first operation is to project these features into the transformer embedding. The linear projection layer flatten the 3 last dimensions and a linear layer is used to learn an embedding of each proposals. This mapping is defined as $\textit{Linear projection}: \mathbb{R}^{B \times (s+1) \times N \times 256 \times 14\times 14}\rightarrow \mathbb{R}^{B \times (s+1)N \times 512}$.
% Transformer encoder decoder

The input of the encoder-decoder transformer is a $(s+1)N$ sequence with $512$ features. The transformer-encoder multi-head attention enables the matching of target frame proposals with respect to the source proposals across time, while the feed-forward learns the matched-motion features. For the transformer-decoder, only the target proposals are used for input. The multi-head attention aggregates the matched-motion features of the encoder to the target proposals and further learns the interactions of the objects by learning an attention between the proposals. 
% predict the pose
Finally a linear layer is used to predict the $6-$DOF pose per object yielding $B\times N \times s \times6$ using a $6$ axis-angle convention parameters. The non-maximum-suppression used for the object detection head is employed to filter the $N=1000$ proposals pose keeping only the relevant objects. The object pose is predicted only for the filtered objects. 

\subsubsection{Ego-pose branch} The ego-pose branch estimates the dominant pose of the camera. The architecture is similar to~\cite{Godard2019}. Since the low-level features that allow matching are usually extracted in the first layers, the $P_4$ features of the FPN for source and target features are used. This network outputs $6$ parameters for the pose transformation using the axis-angle convention.
\subsubsection{Depth branch} The depth branch consists of convolution layers with skip connections from the FPN module as in~\cite{Godard2019}. Similar to prior work~\cite{Zhou2017,Godard2019,Watson2021}, a multi-scale depth is estimated in order to resolve the issue of gradient locality. The prediction of depth at each scale consists of a convolution with a kernel of $1\times1$ and a Sigmoid activation. The output of this activation, $\sigma$, is re-scaled to obtain the depth $D = \frac{1}{a \sigma +b}$, where $a$ and $b$ are chosen to constrain $D$ between $0.5$ and $100$ units, similar to~\cite{Godard2019}.

To maintain self-supervised learning setting, a frozen pretrained EffiecientPS that was trained on the Cityscapes benchmark~\cite{cordts2016cityscapes} is used. This pretrained model achieve $PQ=50.2$ and $SQ=76.8$ on Cityscapes test benchmark. As the representation that was trained for panoptic segmentation may ignore details that are crucial for depth prediction. A duplicate of the Backbone and FPN is used for the Depth and pose heads. This allows learning features optimized for depth prediction without degrading the performances of the panoptic segmentation heads.

\subsection{Objective functions}
Let $\mathcal{L}$ be the objective function. The self-supervised setting casts the depth learning problem into image reconstruction problem through the reverse warping. Thus, learning the parameters $\bm{\theta}$ involves learning $\widehat{\bm{\theta}} \in \Theta$ such $\widehat{\mbf{I}}_t = \mbf{I}_t$. Learning involves minimizing the objective function:
\begin{equation}
    \widehat{\bm{\theta}} = \argmin_{\bm{\theta} \in \Theta} \frac{1}{n} \sum_{n} \mathcal{L} (f(\mbf{I}_t, \widehat{\mbf{I}}_t;\bm{\theta}))
    \label{eq:risk_minimzation}
\end{equation}
where n is the number of training examples. The selected surrogate losses that minimize \refeq{risk_minimzation} are :
\begin{itemize}
    \item \textbf{Photometric loss: } Following~\cite{Zhou2017,Godard2019,Rares2020} The photometric loss seeks to reconstruct the target image by warping the source images using the static/dynamic pose and depth. An $L_1$ loss is defined as follows:
    \begin{equation}
        \mathcal{L}_{rec}(\img{t}, \widehat{\mbf{I}}_{t}) = \sum_{\pt} | \img{t}(\pt) - \widehat{\mbf{I}}_{t}(\pt) |
    \end{equation}
    where $\widehat{\mbf{I}}_{t}(\pt)$ is the reverse warped target image obtained by \refeq{warping}. This simple $L_1$ is regularized using SSIM~\cite{wang2004image} that has a similar objective to reconstruct the image. The final photometric loss is defined as:
    \begin{equation}
    \begin{aligned}
      \mathcal{L}_\textup{pe}(\img{t}, \widehat{\mbf{I}}_{t}) = \sum_{\pt} \big[&(1 - \alpha) \textup{ SSIM}[ \img{t}(\pt) - \widehat{\mbf{I}}_{t}(\pt)]  \\
      & +  \alpha | \img{t}(\pt) - \widehat{\mbf{I}}_{t}(\pt) |\big]
    \end{aligned}
    \label{eq:ssim}
    \end{equation}
    \item \textbf{Depth smoothness: } An edge-aware gradient smoothness constraint is used to regularize the photometric loss. The disparity map is constrained to be locally smooth through the use an image-edge weighted $L_1$ penalty, as discontinuities often occur at image gradients. This regularization is defined as~\cite{heise2013pm}:
    \begin{equation}
    \begin{aligned}
      \mathcal{L}_{s}(D_{t}) = \sum_{p} \big[ &| \partial_x D_{t}(\pt) | e^{-|\partial_x \mbf{I}_{t}(\pt)|}    +\\
      &| \partial_y D_{t}(\pt) | e^{-|\partial_y \mbf{I}_{t}(\pt)|} \big] 
    \end{aligned}\label{eq:loss-disp-smoothness}
\end{equation}
\end{itemize}
The final objective function is defined as :
\begin{equation}
    \mathcal{L} = \mathcal{L}_{pe} + \alpha_d \mathcal{L}_{s}
    \label{eq:objective}
\end{equation}

\begin{table*}
\centering
\footnotesize
\begin{tabular}{|l |cc||c c c c ||c c c|}
\hline
Method & Supervision &Resolution &Abs Rel& Sq Rel&  RMSE & RMSE log  &
$\delta<1.25$ &
$\delta<1.25^2$ &
$\delta<1.25^3$\\
\hline  
SfMlearner~\cite{Zhou2017} & M & 640$\times$192 & 0.183 & 1.595 & 6.709 & 0.270 & 0.734& 0.902 &0.959 \\
GeoNet~\cite{Yin2018}& M+F &416$\times$128  & 0.155 & 1.296 & 5.857  & 0.233 & 0.793 & 0.931 & 0.973\\
CC~\cite{Ranjan2019} & M+S+F& 832$\times$256& 0.140 & 1.070  & 5.326  & 0.217 &0.826 &0.941 &0.975 \\
Self-Mono-SF\cite{hur2020self} & M+F & 832$\times$256 & 0.125 & 0.978 & 4.877   &0.208 & 0.851 & 0.950 & 0.978 \\
Chen \ea~\cite{chen2019towards} & M+S & 512$\times$256   & 0.118 & 0.905 & 5.096 & 0.211 & 0.839 & 0.945 &0.977\\
Monodepth2~\cite{Godard2019} & M & 640$\times$192 & 0.115 & 0.903 & 4.863 & 0.193 & 0.877 & 0.959 & 0.981\\
Lee \ea~\cite{lee2021attentive} & M+F & 832$\times$256 &  0.113 & 0.835 & 4.693 & 0.191 & 0.879 & 0.961 & 0.981 \\
SGDepth~\cite{Klingner2020} & M+S & 1280$\times$384 & 0.113 & 0.835 & 4.693 & 0.191 & 0.879 & 0.961 & 0.981\\
SAFENet~\cite{lou2020safenet}& M+S & 640$\times$192 & 0.112 & 0.788 & 4.582 & 0.187 & 0.878 & 0.963 & 0.983 \\
Insta-DM~\cite{lee2021learning}& M+S & 640$\times$192 & 0.112& 0.777 &4.772 &0.191& 0.872& 0.959& 0.982 \\
PackNetSfm~\cite{Rares2020}& M &640$\times$192 & 0.111 & 0.785 & 4.601 & 0.189 & 0.878 & 0.960 & 0.982\\
MonoDepthSeg~\cite{safadoust2021self} & M & 640$\times$192 & 0.110 & 0.792 & 4.700 & 0.189 & 0.881 & 0.960 & 0.982 \\
Johnston \ea~\cite{johnston2020self} & M & 640$\times$192 & \underline{0.106} & 0.861 & 4.699 & 0.185 & \underline{0.889} & 0.962 & 0.982 \\
Manydepth~\cite{Watson2021}& M+TS& 640$\times$192  & \textbf{0.098} & \underline{0.770} & \textbf{4.459} & \textbf{0.176} & \textbf{0.900 }& \textbf{0.965} & \underline{0.983} \\
\hline
Ours& M+S & 640$\times$192 &    0.110  &   \textbf{0.719}  &   \underline{4.486}  &   \underline{0.184}  &   0.878  &   \underline{0.964}  &   \textbf{0.984 }  \\
\hline
\end{tabular}
\caption{Quantitative performance comparison of on the KITTI benchmark with Eigen split~\cite{Geiger2012CVPR}. For Abs Rel, Sq Rel, RMSE and RMSE log lower is better, and for $\delta < 1.25$, $\delta < 1.25^2$ and $\delta < 1.25^3$ higher is better. The Supervision column illustrates the traing modalities: (M) raw images (S)Semantic, (F) optical flow, (TS) Teacher-student. At test-time, all monocular methods (M) scale the estimated depths with median ground-truth LiDAR.The best scores are bold and the second are underlined}
\label{tab:results}
\end{table*}

\section{Experiment}
\subsection{Setting}
\begin{itemize}
\item{KITTI benchmark~\cite{Geiger2012CVPR}: } 
Following the prior work~\cite{Zhou2017,yang2018unsupervised,Watson2021,Godard2019,Wang2021}, the Eigen~\ea~\cite{Eigen} split is used with Zhou~\ea~\cite{Zhou2017} pre-processing to remove static frames. For evaluation, the metrics of previous works~\cite{Eigen} are used for the depth.

\item{Implementation details: }
PyTorch~\cite{pytorch} is used for all models. The networks are trained for 40 epochs and 20 for the ablation, with a batch size of 2. The Adam optimizer~\cite{kingma2014adam} is used with a learning rate of $lr=10^{-4}$ and $(\beta_1, \beta_2) =  (0.9,0.999)$. Exponential moving average of the model parameters is used with a decay of $decay=0.995$. As the training proceeds, the learning rate is decayed at epoch 15 to $10^{-5}$. The SSIM weight is set to  $\alpha = 0.15$ and the smoothing regularization weight to $\alpha_d = 0.001$. The depth head outputs 4 depth maps. At each scale, the depth is up-scaled to the target image size. The hyperparapmters of EfficientPS are defined in ~\cite{mohan2021efficientps} with $N=1000$ before the Non-maximum-suppression.
Two source images are used $\img{t-1}$ and $\img{t+1}$ are used. The input images are resized to $192 \times 640$. Two data augmentations were performed: horizontal flips with probability $p=0.5$ and color jitter with $p=1$.
\end{itemize}

\subsection{Results}\label{sec:results}
During the evaluation, the depth is capped to 80m. To resolve the scale ambiguity, the predicted depth map is multiplied by the median scaling. The results are reported in~\reftab{results}. The proposed method achieves competitive performances compared to the state-of-the-art (SOTA) and outperforms~\cite{Watson2021} with respect to the $\textit{Sq Rel}$ with an improvement of $6.62\%$. As expected, the proposed method is superior than the prior works that factorize the motion using the optical flow~\cite{Ranjan2019,Yin2018} as their estimated motion is only local and do not account for the class and of the object. Besides, it outperforms other similar methods~\cite{lee2021learning} that factorize using the a pose for each object.~\reffig{qualitative} illustrates the qualitative result comparison. As observed, the proposed method enables high quality depth prediction. Compared to the SOTA methods, our methods is able to represent well the dynamic objects \ie the pedestrians that are crossing the street, the biker on the left. As the network did not mask the dynamic objects during training, the dynamic objects are better learned compared to the methods that masks the dynamic objects~\cite{Watson2021,Rares2020}.   

\subsubsection{Dynamic and static evaluation}
In contrast to training, where the points are categorized into moving and static object's points, testing is performed on all points that have Lidar ground truth. This does not take into account the relevance of the points and the static/dynamic category. Moving objects are crucial for autonomous driving applications. However, with this testing setup, it is not possible to convey how the model performs on moving objects, especially for methods that masks moving objects during training. This begs the question as to whether a model trained with rigid scene assumption learns to represent the depth of dynamic objects even when it is trained with only static objects? 

In order to address this question, the performances of static/dynamic are evaluated separately. A mask of dynamic objects is used to segment moving objects and the assessment can be carried out on each category separately. To avoid biasing the evaluation with the EffiecientPS mask, the evaluation mask is obtained using an independent MaskRCNN~\cite{he2017mask} trained with detectron2~\cite{wu2019detectron2}. The first observation that could be made is that the static objects represents $86.43\%$ of test points. This suggest that using the mean across all points will bias the evaluation towards the static objects. A better solution is to consider the per static/dynamic category mean. \reftab{staticVSdynamic} illustrates the evaluation of the the method versus the current SOTA method video-to-depth prediction~\cite{Watson2021}. The proposed method outperform the SOTA~\cite{Watson2021} for the dynamic objects with a huge gap $\Delta \textit{Sq Rel} = -0.698m$ while the gap for the static objects is only $\Delta \textit{Sq Rel} =+0.011$. The results show that degradation induced by considering the rigid scene assumption is significant. This exposes the limitation of the current evaluation. The KITTI benchmark is biased towards static scene. In order to unbias the evaluation, the mean per-category is used to balance the influence. The proposed method outperform the video-to-depth prediction method~\cite{Watson2021} with $\Delta \textit{Sq Rel} = -0.344m$. The analysis of \reftab{staticVSdynamic} and \reffig{qualitative} suggests that models with rigid scene assumption is still able to predict a depth for moving object (probably due to the depth smoothness regularization and stationary cars) however, its quality is very degraded compared to the static objects. 

Moreover, the results reported in~\reftab{staticVSdynamic} show that the proposed method outperforms Insta-DM~\cite{lee2021learning} with respect to both the static and dynamic objects. Insta-DM~\cite{lee2021learning} proposes an Obj-PoseNet $\mathcal{O}_\psi: \mathbb{R}^{2 \times H \times W \times 3} \rightarrow \mathbb{R}^{6}$ that takes per-object matched binary instance masks $(\mbf{M}_1, \mbf{M}_2)$ and outputs the object pose. It should be noted, however, that the Insta-DM has an unfair advantage since object matching (via binary masks) is provided as input in a supervised learning approach while the proposed method is self-supervised with matching being implicitly learnt in the network. Even so, the proposed method still yields better results on average with respect to dynamic objects. This suggests that the transformer-based instance pose head is able to temporally match the instances and predict an accurate pose.

\subsubsection{Ablation study}

\begin{table}
\begin{center}
\footnotesize
\resizebox{\columnwidth}{!}
{
\begin{tabular}{|c|c|c|c|c|c|}
    \hline
    Evaluation& Model & Abs Rel & Sq Rel & RMSE & RMSE log  \\ 
    \hline
    All points mean &
    \multirow{2}{*}{} 
    ManyDepth \cite{Watson2021} & \textbf{0.098} & \underline{0.770} & \textbf{4.459} & \textbf{0.176}   \\
    &
    Insta-DM~\cite{lee2021learning}& 0.112& 0.777 &4.772 &0.191  \\
    &Ours &  \underline{0.110}  &   \textbf{0.719}  &   \underline{4.486}  &   \underline{0.184} \\
    \hline
    \hline 
    Only dynamic &
    \multirow{2}{*}{} 
    ManyDepth \cite{Watson2021} &   0.192  &   2.609  &   7.461  &   0.288   \\
    &
    Insta-DM~\cite{lee2021learning}& \textbf{0.167}  &   \textbf{1.898}  &   \underline{6.975}  &   \underline{0.283} \\
    &
    Ours & \textbf{0.167}  &   \underline{1.911}  &   \textbf{6.724}  &   \textbf{0.271} \\
    \hline
    \hline 
    Only static &
    \multirow{2}{*}{}
    ManyDepth\cite{Watson2021} &    \textbf{0.085}  &  \textbf{ 0.613 } &   \textbf{4.128}  &   \textbf{0.150}    \\ &
    Insta-DM~\cite{lee2021learning} &  0.106  &   0.701  &   4.569  &   0.171   \\
    &
    Ours & \underline{0.101}  &   \underline{0.624}  &   \underline{4.269}  &   \underline{0.163} \\
    \hline
    Per category mean & \multirow{2}{*}{}
    ManyDepth\cite{Watson2021} &   0.139   &  1.611 & 5.794   &  \underline{0.219} \\ &
    Insta-DM~\cite{lee2021learning}  &   \underline{0.137}   &  \underline{1.299} &  \underline{5.772}  &  0.227\\ &
    Ours & \textbf{0.134}  &  \textbf{1,267}  &   \textbf{5,496}& \textbf{0,217}  \\
    \hline
    \end{tabular}
}
\caption{Quantitative performance comparison for dynamic and static objects. The proposed method outperforms the SOTA~\cite{Watson2021} that uses masking for the dynamic objects with a huge gap $\Delta \textit{Sq Rel} = -0.698m$. In addition, it outperforms Insta-DM~\cite{lee2021learning} which explicitly models dynamic objects.}
\label{tab:staticVSdynamic}

\end{center}
\end{table}
\begin{figure*}
    \centering
    \includegraphics[width=\textwidth]{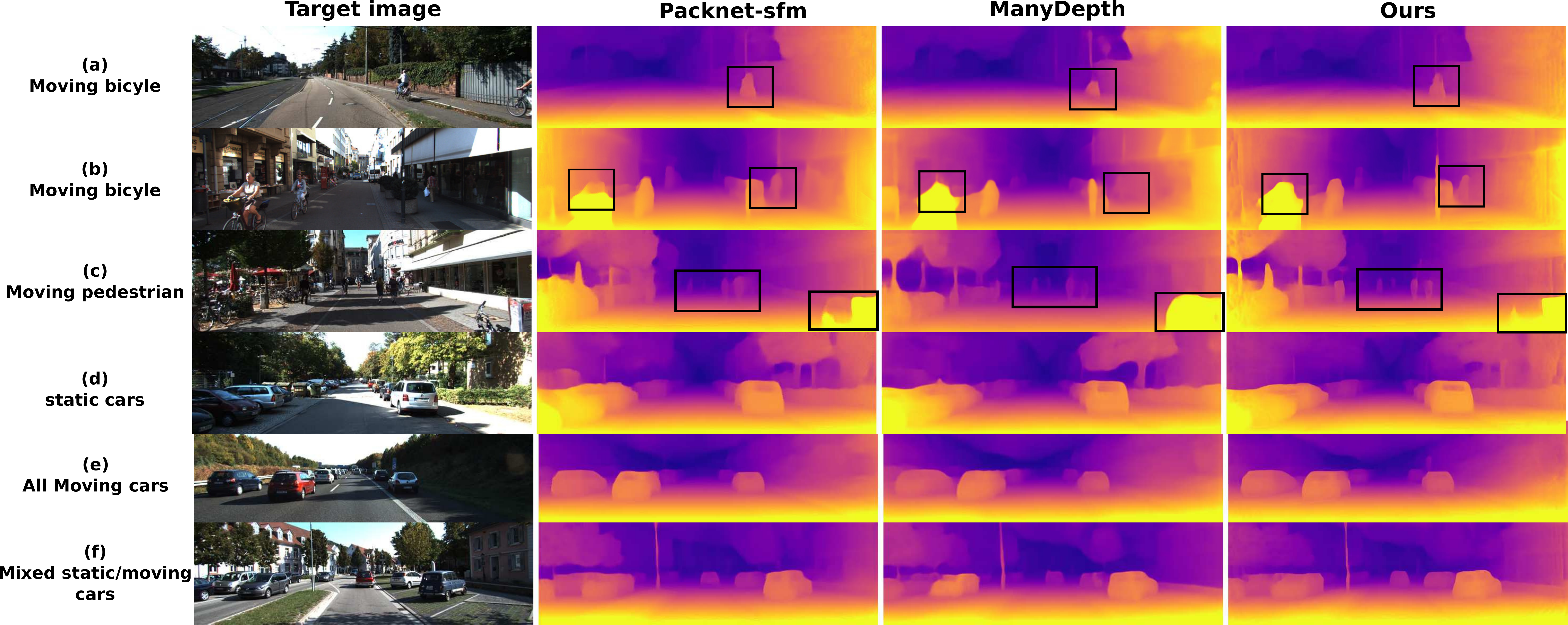}
    \caption{Qualitative results of the proposed method with SOTA methods~\cite{Watson2021,Rares2020}. (a) The proposed method is on par with the baselines for static objects. (e) and (f) show cars as moving objects. Although the baselines are trained with auto-masking, the dataset is rich with static cars that are not masked during training, this provides clues to learn the depth for moving cars. (d-g) show more complex situations as pedestrians and bicycles tend to always move in the KITTI dataset. The qualitative results show that the proposed method outperforms the baselines. This is validated further by the quantitative results reported in \reftab{staticVSdynamic}}
    \label{fig:qualitative_2}
\end{figure*}
%% Motivate the use of the 
\begin{table*}
\begin{center}
\footnotesize
\begin{tabular}{|c||c|c|c|c||c|c|c|c|}
\hline
Ablation & Backbone & Ego-pose input feature & Shared backbone & Piece-wise rigid pose  &Abs Rel & Sq Rel & RMSE & RMSE log  \\ 
\hline
\hline
A1& Resnet18~\cite{he2016deep} & Layer5 & - & - &	 0.121  &   0.914  &   4.890  &   0.196 \\% &   0.866  &   0.958  &   0.981  \\
\hline
\hline
A2 &EfficientNet-b5 & $P_{16}$ & - & - &	  0.132  &   0.906  &   4.981  &   0.205 \\%	&	0.877	&	0.981 \\
A3&EfficientNet-b5 & $P_8$ & - & - &	  0.127  &   0.983  &   5.010  &   0.201 \\%	&	0.877	&	0.981 \\
A4& EfficientNet-b5 & $P_4$ & - & - &   0.121  &   0.894  &   4.886  &   0.197 \\% &   0.864  &   0.957  &   0.981  \\
A5& EfficientNet-b5 &  $P_4$ &  \checkmark & - &   0.120  &   0.925  &   4.868  &   0.194 \\% &   0.864  &   0.958  &   0.982  \\
\hline
\hline
A6& EfficientNet-b5 &  $P_4$ & \checkmark & \checkmark &   0.113  &   0.795  &   4.689  &   0.190  \\%&   0.871  &   0.960  &   0.982  \\
A7& EfficientNet-b6 & $P_4$ & \checkmark & \checkmark &  0.110  &   0.719  &   4.486  &   0.184  \\%&   0.878  &   0.964  &   0.984   \\
\hline
\end{tabular}
\caption{An ablation study of the proposed method. The evaluation was done on KITTI benchmark using Eigen split~\cite{Eigen}. As observed, The effect of the backbone is minimal A1 vs A5, The choice of the input feature for ego-pose head is sensible A2 vs A3 vs A4, The performance of the proposed method is obtained mainly by the introduction of the piece-wise rigid pose A5 vs A6. Increasing the complexity of the model allow better performances and better training stability A6 vs A7}
\label{tab:ablations}
\end{center}
\end{table*}

\reftab{ablations} illustrate an ablation study performed to validate the contribution of the proposed method. The results strongly suggest that the performance of the proposed network is mainly obtained by the introduction of the motion factorization through the proposed instance pose head. %The first observation that could be made is that the effect of the backbone (A1 versus A4) and sharing the backbone (A4 versus A5) is minor.

\begin{itemize}
    \item \textbf{A1 versus A4:} Introducing a more complex architecture did not contribute to the improvement of the performances.
    \item \textbf{A4 versus A5:} Sharing the backbone for the depth network did not contribute to the improvement of the performances. However, it did reduce the complexity of the network.
    \item \textbf{A5 versus A6: } Introducing the piece-wise rigid pose warping induces an improvement of $\Delta \textit{Sq rel}=14.1\%$
    \item \textbf{A2 versus A3 versus A4: } The pose head is sensitive to the choice of the features level. $P_4$ is the optimal level for this application.
\end{itemize}

This results suggest that not only the models learn an accurate depth but also accurate instance pose. This result demonstrates that the transformer network is able to match and learn the interaction of the objects across time. The model in $A5$ is on the same setting of the other SOTA methods~\cite{Watson2021,Rares2020}. Despite using this low performance baseline, the introduction of the dynamic warping enabled the proposed method to achieve the SOTA results. 

An interesting observation during training is that $A6$ under-fits the data (i.e., the validation loss is less than the learning loss). The test performances are not stable, the best model among the 20 epochs is reported for this backbone. In order to resolve this under-fitting, the complexity of the model is increased $A7$. This allow for a better stability of the training loss and test performance. The best results are obtained using this complexity.
The additional instance pose results in an additional run-time overhead during training. The training time for 1 epoch for A5 and A7 is $233mn$ and $58mn$ trained on RTX3090 respectively. However, the additional run-time is only for the training. At test-time, the depth network require only a single pass of the image $\img{t}$ with roughly 34FPS for $A7$ model and 38FPS for $A6$ model using a single RTX3090.

\section{Conclusion}

In this paper a novel instance pose head is introduced for self-supervising monocular depth prediction. This head enables the factorization of the scene's motion. Thus, alleviating the rigid scene assumption. It is shown that it achieves the SOTA results on the KITTI benchmark~\cite{Geiger2012CVPR}. The ablation study further validates that the multi-head attention of the transformer network predicts an accurate object pose. Moreover, the impact of the dynamic motion on this benchmark is exposed. Namely, the bias towards static objects where $86.43\%$ of the test pixels correspond to static objects. A mean per static/dynamic category metric is proposed to unbias the assessment. 
%Future work will consider extending an object tracker with the object pose that explicitly optimized for matching.  

\bibliographystyle{plain}
\bibliography{biblio}

\end{document}